\documentclass{article}
\usepackage{spconf,amsmath,graphicx}
\usepackage{hyperref}

\usepackage{arydshln}
\usepackage{multirow}
\usepackage{amssymb}
\usepackage{booktabs}

\usepackage{xcolor}

\title{``How Robust r u?'':\\ Evaluating Task-Oriented Dialogue Systems on Spoken Conversations}
\name{\begin{tabular}{c}Seokhwan Kim, Yang Liu, Di Jin, Alexandros Papangelis,\\Karthik Gopalakrishnan, Behnam Hedayatnia, Dilek Hakkani-T\"ur\end{tabular}}
\address{Amazon Alexa AI, Sunnyvale, CA, USA\\
}

\begin{document}
%
\maketitle
\begin{abstract}
  Most prior work in dialogue modeling has been on written conversations mostly because of existing data sets.
  However, written dialogues are not sufficient to fully capture the nature of spoken conversations as well as the potential speech recognition errors in practical spoken dialogue systems.
  This work presents a new benchmark on spoken task-oriented conversations, which is intended to study multi-domain dialogue state tracking and knowledge-grounded dialogue modeling.
  We report that the existing state-of-the-art models trained on written conversations are not performing well on our spoken data, as expected.
  Furthermore, we observe improvements in task performances when leveraging $n$-best speech recognition hypotheses such as by combining predictions based on individual hypotheses.   
  Our data set enables speech-based benchmarking of task-oriented dialogue systems.
\end{abstract}
\begin{keywords}
spoken dialogue systems, dialogue state tracking, knowledge-grounded dialogue generation
\end{keywords}

\section{Introduction}
\label{sec:intro}

Recently, more public data sets and benchmarks have become available for dialogue research on task-oriented conversations in various domains~\cite{el2017frames,wen2017network,budzianowski2018multiwoz,rastogi2020towards}. 
However, most data sets include only written conversations collected by crowdsourcing via web interfaces, which differ from spoken conversations for the following reasons.
First, there are differences between the style of spoken and written conversations, even for the same context, intention, and semantics.
Second, spoken conversations tend to have extra noise from grammatical errors, disfluencies or barge-ins, which are rarely encountered when processing written texts.
In addition, speech recognition errors bring about even more challenges for developing spoken dialogue systems in practice.

Figure~\ref{fig:written-spoken} compares a written and a spoken conversation, and shows many differences in terms of wording and expressions between the two examples even for the same content.
The spoken example includes disfluencies and speech recognition errors highlighted as underlined text.
Moreover, no punctuation, capitalization, or sentence segmentation is available in raw speech recognizer outputs.

\begin{figure}[t]
\footnotesize
\begin{tabular}{l p{7cm}}
  \multicolumn{2}{l}{\textbf{Written Conversation}} \\ \hline
  User & I need a hotel in Fisherman's Wharf \\
  Agent & Is there a particular price range you are looking for? \\
  User & I'm looking in the expensive price range \\
  Agent & The Suite at Fisherman's Wharf may work for you \\
  User & Do you know how much the parking is? \\
  Agent & It would cost 25 dollars per day. \\ 
  \\
  \multicolumn{2}{l}{\textbf{Spoken Conversation}} \\ \hline
  User & hi \underline{ummm} i’m looking for a place at \underline{uhhh} to stay at fisherman’s wharf at a hotel in the expensive \underline{pressure engine} \\
  Agent & sure let me see ok so there is one called the suites at fisherman's wharf is that something that would be interesting to you \\
  User & can you tell me how much parking \underline{coast} \\
  Agent & sure okay this hotel charges twenty five dollars per day\\
\end{tabular}
\caption{Examples of written and spoken conversations}
\label{fig:written-spoken}
\end{figure}

There have been extensive studies towards robust language understanding against spoken input, mostly for single-turn intent classification and slot filling tasks~\cite{tur2002improving,hakkani2006beyond,6424218,tur2013semantic,masumura2018neural,Ladhak2016,velikovich2016semantic,huang2019adapting,huang-chen-2020-learning}. Nonetheless, the research communities have rarely addressed these issues on more contextual dialogue tasks including dialogue state tracking, dialogue policy learning, or end-to-end dialogue generation, which are as important as the single-turn understanding tasks in fully working dialogue systems.
This is mainly due to the lack of rich, annotated spoken data for such multi-turn dialogue tasks.

To benchmark the robustness of conversational models on spoken conversations, this work introduces a new data set with spoken task-oriented dialogues.
We release this data as the official validation set of our public benchmark challenge under DSTC10~\footnote{https://dstc10.dstc.community/} for the following two subtasks:
1) multi-domain dialogue state tracking~\cite{budzianowski2018multiwoz}
and 2) knowledge-grounded dialogue modeling~\cite{kim2020beyond}.
The remainder of this paper presents the data and task details as well as the analysis result showing how state-of-the-art methods from existing written data perform on these benchmark tasks.

\section{Related Work}
\label{sec:related_work}
There has been a lot of studies towards improving the single-turn spoken language understanding (SLU) robustness to automatic speech recognition (ASR) errors.
The most common line of work has focused on utilizing multiple recognition hypotheses given from ASR systems in the form of word confusion networks~\cite{tur2002improving,hakkani2006beyond,6424218,tur2013semantic,masumura2018neural} or word lattices~\cite{Ladhak2016,velikovich2016semantic,huang2019adapting,huang-chen-2020-learning}.
Recently, more proactive approaches have been explored for ASR error correction~\cite{9053213,namazifar2021correcting} and data augmentation~\cite{wang2020data} as well.

The earlier dialog state tracking challenges (DSTCs) aimed to address the multi-turn dialogue problems in spoken conversations.
DSTC2~\cite{henderson-etal-2014-second} and DSTC3~\cite{7078595} data sets included $n$-best ASR hypotheses as well as word confusions, which was intended for speech-oriented studies.
However, the dialogue research community hasn't paid much attention to this aspect, due to the lack of critical ASR issues on these single-domain human-machine conversations that were restricted only to a small domain ontology.
DSTC4~\cite{kim2017fourth} and DSTC5~\cite{7846311} targeted to extend it with multi-domain human-human conversations, but only manual transcriptions were included in the challenge data sets without ASR outputs.

On the other hand, many recent task-oriented dialogue data sets~\cite{el2017frames,wen2017network,budzianowski2018multiwoz,rastogi2020towards} include written conversations collected by crowdsourcing with no consideration of speech-specific aspects in spoken dialogue systems.
There have been studies focusing on the speech robustness issues on task-oriented~\cite{peng2020raddle} and open-domain conversations~\cite{gopalakrishnan2020neural}, but they were restricted to simulated ASR errors on top of written conversations.

\section{Data}
\label{sec:data}

To study speech-based task-oriented dialogue modeling, we collected spoken human-human dialogues about touristic information for San Francisco.
Each session was collected by pairing two participants: one as a user and the other as an agent.
We provided a set of specific goals to the user-side participant before each session.
The agent-side participant had access to the domain database including both structured information and unstructured text snippets. 
We recorded 107 sessions, which are around 4 hours in total, and manually transcribed all the utterances.
This data is released as the official validation set of our public benchmark challenge under DSTC10 Track 2. 

Table~\ref{tbl:data_stats} summarizes the data details in comparison with two other data sets.
MultiWOZ 2.0~\cite{budzianowski2018multiwoz} and its variants~\cite{eric2019multiwoz,zang2020multiwoz} include crowd-sourced written conversations about seven different domains including hotel, restaurant, attraction, train, taxi, hospital, and police station in Cambridge, UK.
Following the MultiWOZ data collection set-up, we recently released new written dialogues as a part of the official test set for the DSTC9 Track 1~\cite{kim2021domain}.
This data was collected for a new locale, San Francisco, for three target domains: hotel, restaurant, and attraction, but with almost three times more entities than the MultiWOZ ontology entries.
In addition, this data includes the turns grounded on the knowledge snippets from the FAQ list for the entities.
All these infrastructure and domain ontology for this written data were re-used for our spoken data collection.
The difference is that the DSTC10 data came from the recorded conversations instead of the written texts from crowd-sourcing. 

\begin{table*}
  \small
  \centering
  \caption{Comparisons of the data sets.}
  \label{tbl:data_stats}
  \begin{tabular}{l l l l r r r r r}
    & & & & \multicolumn{2}{c}{Dialogues} & \multicolumn{3}{c}{Domain Ontology}\\
    Data & Split & Locale & Modality & \# sessions & \# turns & \# domains & \# entities & \# snippets \\ \hline
    MultiWOZ 2.0 & all & Cambridge & written & 8,438 & 113,556 & 7 & 289 & - \\ 
    DSTC9: Track 1 & test & San Francisco & written & 903 & 8,501 & 3 & 855 & 15,086\\ \hdashline[.4pt/1pt]
    DSTC10: Track 2 & val & San Francisco & spoken & 107 & 2,292 & 3 & 855 & 15,086\\
  \end{tabular} 
\end{table*}

To benchmark the robustness of models in practical spoken dialogues systems, we test on the ASR output instead of manual transcript for each user turn.
In terms of the ASR model, we took the wav2vec 2.0~\cite{baevski2020wav2vec} model pre-trained on 960 hours of Librispeech~\cite{panayotov2015librispeech} and fine-tuned it with 10\% of our data.
Then, we run 10-best decoding with an external language model built with KenLM~\cite{heafield2011kenlm} on all the written texts from MultiWOZ and DSTC9 data sets.
This ASR pipeline finally achieved a 24.09\% WER at 1-best and 21.89\% oracle WER at 10-best hypotheses on the user utterances of the remaining 90\% of our data.


\begin{figure*}
\small
\begin{tabular}{l p{8cm} c p{6cm}}
Speaker & Utterance & Task & Ground-truth Annotations \\ \hline
  User & hi ummm i’m looking for a place at uhhh to stay at fisherman’s wharf at a hotel in the moderate pressure engine & Task 1 & hotel-area: fisherman's wharf \leavevmode\newline hotel-type: hotel \leavevmode\newline hotel-pricerange: moderate \\
  Agent & sure let me see what can i find for you ok unfortunately we're not showing any result is there any specification that i can change to possibly find you something \\ \hdashline[.4pt/1pt]
  User & is there wholesale in the expensive pressure engine student & Task 1 & hotel-area: fisherman's wharf \newline hotel-type: hotel \newline hotel-pricerange: expensive \\
  Agent & sure let me see ok so there is one called the suites at fisherman's wharf is that something that would be interesting to you\\ \hdashline[.4pt/1pt]
  User & can you tell me how much parking coast & Task 2 & Q: What is the parking cost at the Suites at Fisherman's Wharf? - A: Parking costs \$25 per day.\\
  Agent & sure okay this hotel charges twenty five dollars per day\\
\end{tabular}
\caption{An example conversation with the ground-truth labels for both tasks.} 
\label{fig:example}
\end{figure*}

\section{Tasks}
\label{sec:tasks}

\begin{figure*}[t]
\centering
\includegraphics[width=\linewidth]{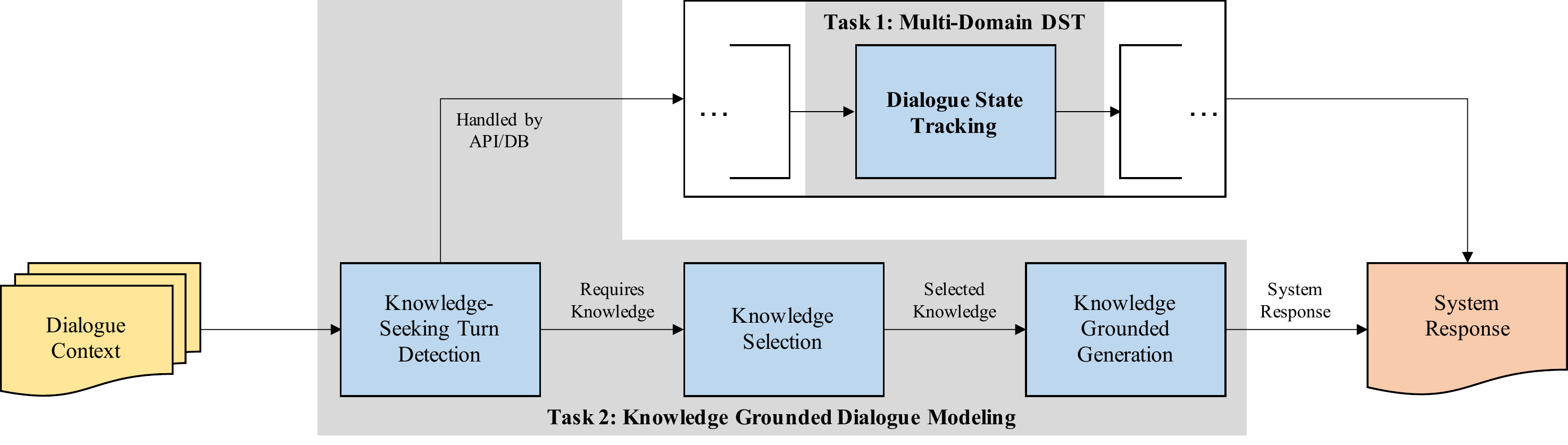}
\caption{An overview of the benchmark tasks: multi-domain dialogue state tracking and knowledge-grounded dialogue modeling.}
\label{fig:overall}
\end{figure*}

This section describes two benchmark tasks that we propose in this work. 
As shown in Figure~\ref{fig:overall}, we decouple between turns that could be handled by conventional task-oriented conversational models with no extra knowledge and turns that require external knowledge resources, following the architecture in ~\cite{kim2020beyond}.
In the first API/DB-based pipeline, we focus only on dialogue state tracking as the first target task.
For the other knowledge access branch, our task 2 includes all three subtasks: 1) Knowledge-seeking Turn Detection, 2) Knowledge Selection, and 3) Knowledge-grounded Response Generation introduced in ~\cite{kim2020beyond}.


\subsection{Task 1: Multi-domain Dialogue State Tracking}
\label{sec:tasks:dst}

Dialogue state tracking (DST) aims to estimate the system's belief states after each interaction with the user, which is a key problem in task-oriented conversational modeling.
The belief states are defined to represent the latest user goals in a dialogue context from the beginning to the target user turn of a given conversation.

In this benchmark, we address the multi-domain DST task on human-human conversations, which have been actively explored by the dialogue research community mainly with MultiWOZ and its variants~\cite{budzianowski2018multiwoz,eric2019multiwoz,zang2020multiwoz}.
Following previous work, we also represent the user goals as a set of slot-value pairs defined for each domain and take the slot-level value prediction performances and the joint goal accuracy~\cite{henderson-etal-2014-second} as the evaluation metrics.
Figure~\ref{fig:example} presents an example conversation with the ground-truth DST annotations for the first two user turns.

Our task differs from most previous DST benchmarks in the following two aspects.
First, we focus on the DST performances on spoken conversations rather than written ones. 
The latter has been widely used by previous DST studies because of its cost-efficiency in large scale data collection.
We believe that those written data sets are not enough to fully reflect the actual human behaviors for spoken conversations.
So we propose to take DST models trained on existing written data,
evaluate on our new collection of spoken conversations,
and eventually try to improve the model performance in face of a mismatch between training and test data sets.

In addition, our new data uses the ASR output instead of manual transcripts for the user turns, as described in Section~\ref{sec:data}.
The goal is to evaluate how robust each DST model is against ASR errors.
Although ASR errors are expected to be a very critical issue in developing spoken dialogue systems in practice,
it hasn't been studied for the DST task. 


\subsection{Task 2: Knowledge-grounded Dialogue Modeling}
\label{sec:tasks:kmdm}

Recently, we introduced a new benchmark on task-oriented conversational modeling with unstructured knowledge access~\cite{kim2020beyond}, which aims to incorporate external unstructured knowledge into the end-to-end dialogue response generation problem.
As shown in Figure~\ref{fig:overall}, this task focuses on the knowledge access branch, including the following three sub-tasks:
\begin{itemize}
\item \textbf{Knowledge-seeking Turn Detection} decides whether to trigger the knowledge access branch for a given utterance and dialogue history
\item \textbf{Knowledge Selection} selects proper knowledge snippets from the domain knowledge base for the knowledge seeking turn
\item \textbf{Knowledge-grounded Response Generation} generates a system response given a triple of input utterance, dialog context, and the selected knowledge snippet
\end{itemize}

Figure~\ref{fig:example} shows an example knowledge-seeking turn along with its ground-truth knowledge snippet and a reference response. 
We organized a challenge track on this task under DSTC9 \cite{kim2021domain,gunasekara2020overview}, which had more than 100 submissions from 24 teams in total.

In this work, we propose to extend this DSTC9 track by replacing written conversations with spoken ones, as in the first task (Section~\ref{sec:tasks:dst}).
The issue between written and spoken conversations has been partially discussed in our DSTC9 track with a spoken subset in the test data~\cite{kim2021domain}.
But that was only on manual transcripts, while the new data for this work includes the ASR outputs for the user turns.

\section{Baseline Models}
\label{sec:models}

To investigate the existing model behaviors on our spoken data, we took state-of-the-art models on written data for both tasks as baselines.

\subsection{Task 1 Baseline}
\label{sec:models:task1}

We use TripPy~\footnote{https://gitlab.cs.uni-duesseldorf.de/general/dsml/trippy-public}~\cite{heck2020trippy} as a baseline for the DST task.
It is based on a fine-tuned BERT~\cite{devlin2019bert} on the DST objectives with copy mechanisms from three different sources: user utterance, system utterance, and previous dialogue states.
This model achieved 55.30\% in joint goal accuracy on MultiWOZ 2.1, which is also used as the DST baseline for DialoGLUE~\cite{mehri2020dialoglue}.



\subsection{Task 2 Baselines}
\label{sec:models:task2}

For task 2, we take two baselines: one is the official baseline of DSTC9~\footnote{https://github.com/alexa/alexa-with-dstc9-track1-dataset}~\cite{kim2021domain} and the other is Knover~\footnote{https://github.com/PaddlePaddle/Knover}~\cite{he2021learning} from the DSTC9 track winner.
The official DSTC9 baseline uses the fine-tuned GPT-2~\cite{radford2019language} for each sub-task.
For Knover, we use the model for their submission \#0.
It has the following three updates from the baseline: 1) replacing GPT-2 with PLATO-2~\cite{bao2021plato2}; 2) multi-scale negative sampling for knowledge selection; and 3) response generation with beam search instead of nucleus sampling.
Note that the DSTC9 winning entry was a further enhanced version from this model by schema guided turn detection and model ensemble, however, that model is not publicly available.

\section{Evaluation}
\label{sec:evaluation}

\subsection{Task 1 Results}
\label{sec:evaluation:task1}

\begin{table*}[t]
  \small
  \centering
  \caption{Multi-domain dialogue state tracking results by the TripPy baseline on the DSTC10 validation and MultiWOZ 2.1 test data sets.}
  \label{tbl:dst_results}
  \begin{tabular}{l c c c c c c c c c}
    & Joint Goal& Slot & \multicolumn{3}{c}{Value Prediction} & \multicolumn{3}{c}{None Prediction}\\
    Data & Accuracy & Accuracy & Precision & Recall & F1 & Precision & Recall & F1\\ \hline
    DSTC10 (1-best) & 0.0043 & 0.7208 & 0.5977 & 0.3378 & 0.4317 & 0.7473 & 0.9788 & 0.8475 \\
    \hdashline[.4pt/1pt]
    MultiWOZ & 0.5427 & 0.9738 & 0.9056 & 0.9103 & 0.9080 & 0.9802 & 0.9934 & 0.9868\\
  \end{tabular}
\end{table*}

Table~\ref{tbl:dst_results} compares the TripPy baseline performances on the DSTC10 and MultiWOZ 2.1 data sets.
Note that all the evaluations on the DSTC10 data were done on the official task 1 validation set including 936 target turns. 
It achieves extremely low performance on the spoken data in joint goal accuracy and also a significantly worse score in slot-level accuracy compared to that on MultiWOZ.
Considering that there are too many false negatives from the model predictions, we report more detailed results in precision, recall and F1 scores for {\it none} predictions and other values from the system separately.
The results show that the baseline model is not working well on our test data, especially in value predictions.

\begin{figure}[t]
\centering
\includegraphics[width=\linewidth]{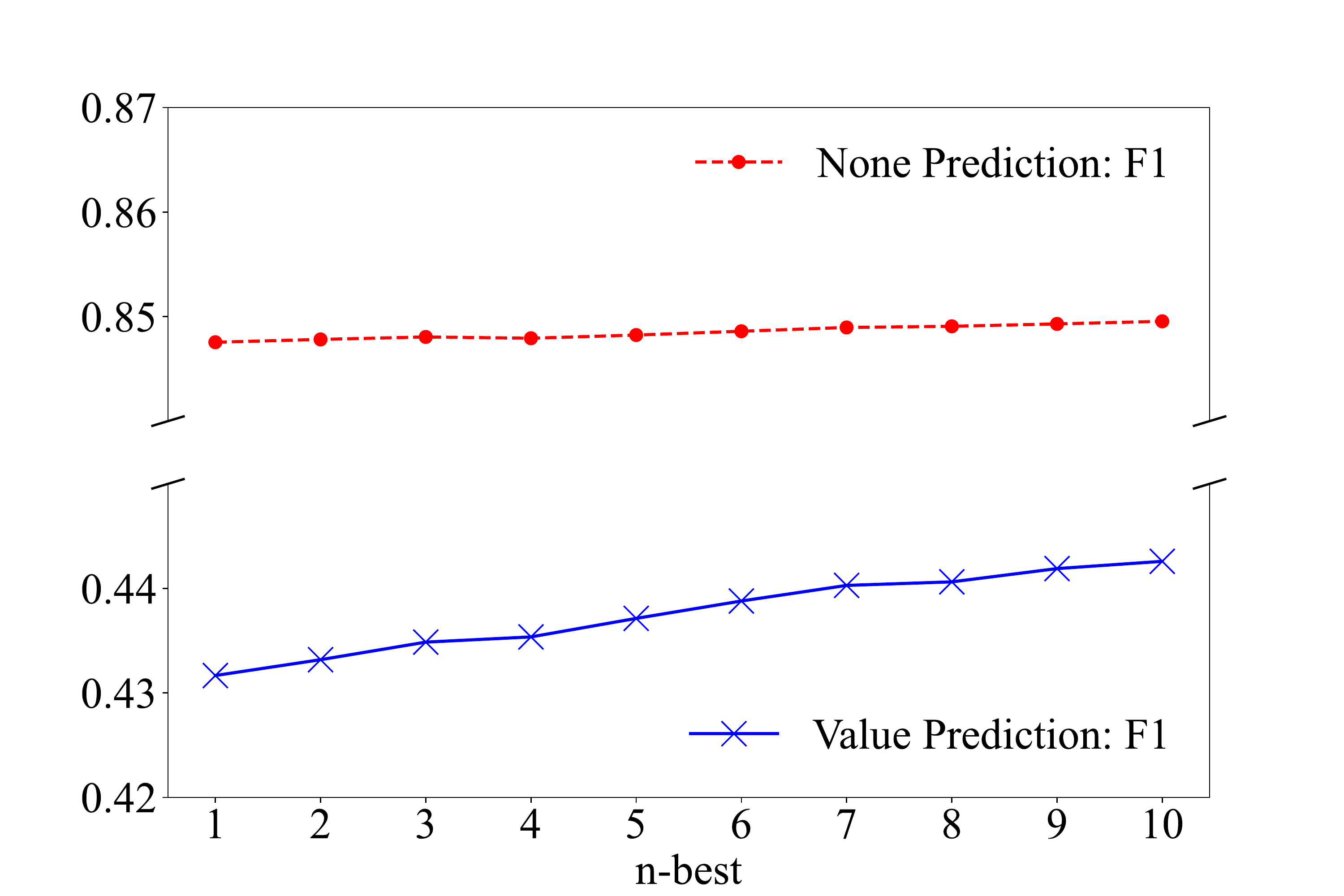}
\caption{Value and none prediction performances with $n$-best ASR hypotheses}
\label{fig:dst_nbest}
\end{figure}

To utilize $n$-best ASR results, we first took every combination of the ASR hypotheses of the user turns and calculated the sum of the turn-level language model scores to select the top-$n$ context-level hypotheses among them.
Then, we run the TripPy model for each context separately and finally aggregated all the predicted values by taking the union of the $n$ results for each slot.
Figure~\ref{fig:dst_nbest} presents the value and none prediction performances with different $n$ values.
The use of multiple ASR hypotheses helped to achieve slightly higher scores than using only the top-1 results, but the differences were not significant.


\begin{figure}[t]
\centering
\includegraphics[width=\linewidth]{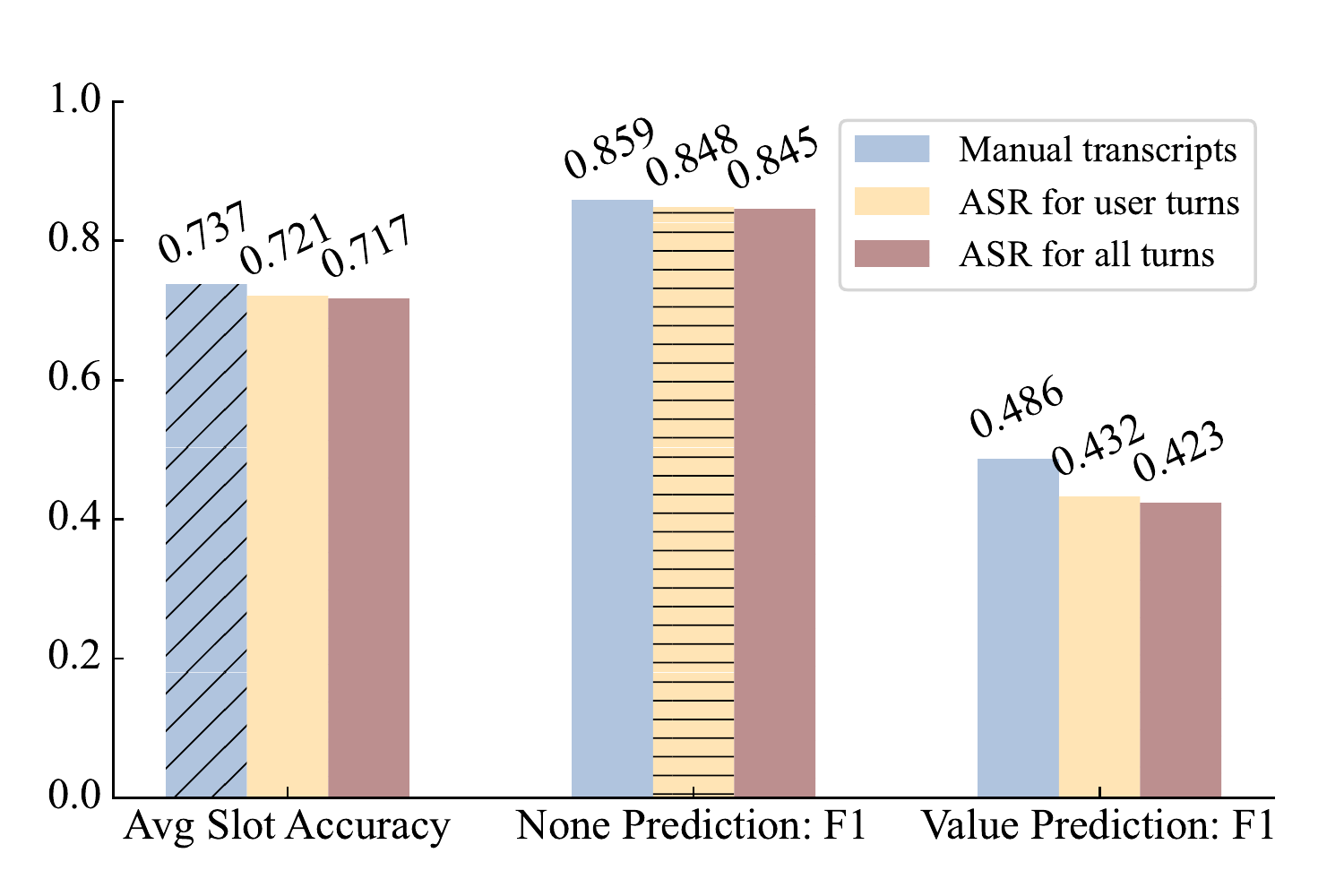}
\caption{Comparisons of the task 1 baseline model performances on manual transcripts and ASR results}
\label{fig:dst_modes}
\end{figure}

To further investigate how much ASR errors affect DST performance, we run the model on two additional configurations: one with manual transcripts with no ASR errors, and the other with the 1-best ASR results not only for the user turns, but also for the agent turns.
Figure~\ref{fig:dst_modes} compares the model performance on those three different settings and shows the impact of the ASR errors on all three metrics.
In particular, the value prediction score in F1 drops 5\% and 6\%, when the ASR errors are introduced in the user turns and both turns, respectively.
But even with no ASR error, the value prediction score at an F1 of 48.61\% is much lower than 90.80\% on the original MultiWOZ.
This indicates that the other aspects including different modalities (spoken vs written) and locales (Cambridge vs San Francisco) between training and test data are as critical as the ASR errors, and they all result in performance degradation.

\subsection{Task 2 Results}
\label{sec:evaluation:task2}

\begin{table}[t]
  \small
  \centering
  \caption{Statistics of the task 2 data sets. $\dagger$ indicates the spoken data with manual transcriptions only.}
  \label{tbl:task2_data_stats}
  \begin{tabular}{l l l l r r r}
    & & & & \# & total \# \\
    Dataset & Split & Locale & Modality & dialogs & instances \\ \hline
    DSTC9 & Test & CAM & written & 977 & 2,084 \\
    & Test & SF & written & 900 & 1,834 \\
    & Test & SF & spoken$^\dagger$ & 107 & 263 \\ \hdashline[.4pt/1pt]
    DSTC10 & Valid & SF & spoken & 107 & 263 \\
  \end{tabular}
\end{table}

This section reports the task 2 baseline performances on both the DSTC9 test set and DSTC10 validation set (data statistics are shown in Table~\ref{tbl:task2_data_stats}). 
We treat each instance including a target turn and its dialogue context independently from the others in the data and evaluated the prediction results at the point of the final target turn only.
Table~\ref{tbl:detection_selection_results} shows the knowledge-seeking turn detection and knowledge selection performance of the baseline models.
For both tasks, the models achieve significantly worse performance on the spoken DSTC10 data compared to the whole DSTC9 test set.
We notice there are many false negatives in the turn detection task, which was the key factor of the performance degradation due to the low recalls.
Comparing these two baseline models on the spoken data, the DSTC9 baseline method achieves slightly better scores for the detection task, while Knover is better for two of the three selection metrics (MRR@5 and Recall@1).
The response generation results in Table~\ref{tbl:generation_results} also show the same pattern, with lower scores than the ones on the DSTC9 test set.
While Knover consistently outperformed the baseline on the DSTC9 test set for all the metrics, there was no dominant result by either system on the DSTC10 validation data.

\begin{table*}[t]
  \small
  \centering
  \caption{Knowledge-seeking turn detection and knowledge selection results by the DSTC9 and Knover baselines on the DSTC10 validation and DSTC9 test data sets.}
  \label{tbl:detection_selection_results}
  \begin{tabular}{l l c c c c c c c c}
    & & & \multicolumn{3}{c}{Detection} & & \multicolumn{3}{c}{Selection} \\
    Data & Model & & Precision & Recall & F1 & & MRR@5 & Recall@1 & Recall@5 \\ \hline
    DSTC10 & Baseline & & \bf{0.9851} & \bf{0.6346} & \bf{0.7719} & & 0.5405 & 0.4444 & 0.6901 \\
           & Knover & & 0.9701 & 0.6250 & 0.7602 & & \bf{0.5782} & \bf{0.5263} & \bf{0.6667} \\ \hdashline[.4pt/1pt]
    DSTC9 & Baseline & & 0.9933 & 0.9021 & 0.9455 & & 0.7263 & 0.6201 & 0.8772\\
         & Knover & & 0.9941 & 0.9430 & 0.9679 & & 0.9181 & 0.8870 & 0.9554\\
  \end{tabular}
\end{table*}

\begin{table*}[t]
  \small
  \centering
  \caption{Response generation results in automated metrics on the DSTC10 validation and DSTC9 test data sets.}
  \label{tbl:generation_results}
  \begin{tabular}{l l l c c c c c c c c}
    Data & Model & BLEU-1 & BLEU-2 & BLEU-3 & BLEU-4 & METEOR & ROUGE-1 & ROUGE-2 & ROUGE-L \\ \hline
    DSTC10 & Baseline & \bf{0.1338} & 0.0539 & 0.0235 & \bf{0.0128} & 0.1457 & \bf{0.1650} & 0.0443 & 0.1181 \\
           & Knover & 0.1301 & \bf{0.0635} & \bf{0.0330} & 0.0111 & \bf{0.1592} & 0.1578 & \bf{0.0536} & \bf{0.1245} \\ \hdashline[.4pt/1pt]
    DSTC9 & Baseline & 0.3031 & 0.1732 & 0.1005 & 0.0655 & 0.2983 & 0.3386 & 0.1364 & 0.3039 \\
         & Knover & 0.3726 & 0.2402 & 0.1556 & 0.1064 & 0.3802 & 0.4103 & 0.1936 & 0.3665 \\
  \end{tabular}
\end{table*}

\begin{figure}[t]
\centering
\includegraphics[width=\linewidth]{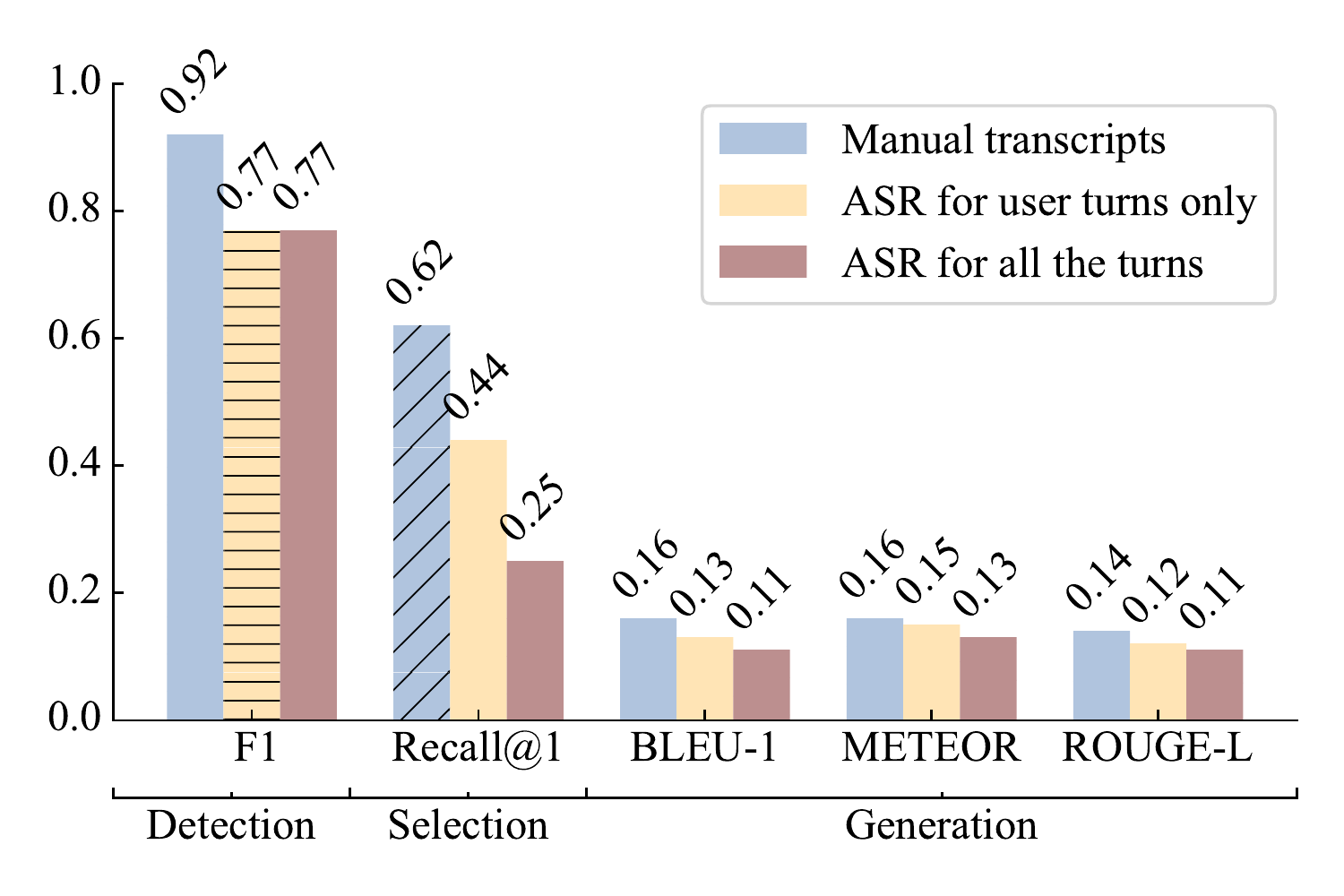}
\caption{Comparisons of the task 2 baseline performances on manual transcripts and ASR results}
\label{fig:kmdm_modes}
\end{figure}

Figure~\ref{fig:kmdm_modes} shows the baseline model performance in different configurations: with manual transcripts, and 1-best ASR outputs only for user turns or for all the turns.
We can see that the ASR errors have a much bigger impact on these tasks compared to the same experimental results on the DST task.
There is a larger degradation in the detection and selection performances in face of ASR errors.

\begin{figure}[t]
\centering
\includegraphics[width=\linewidth]{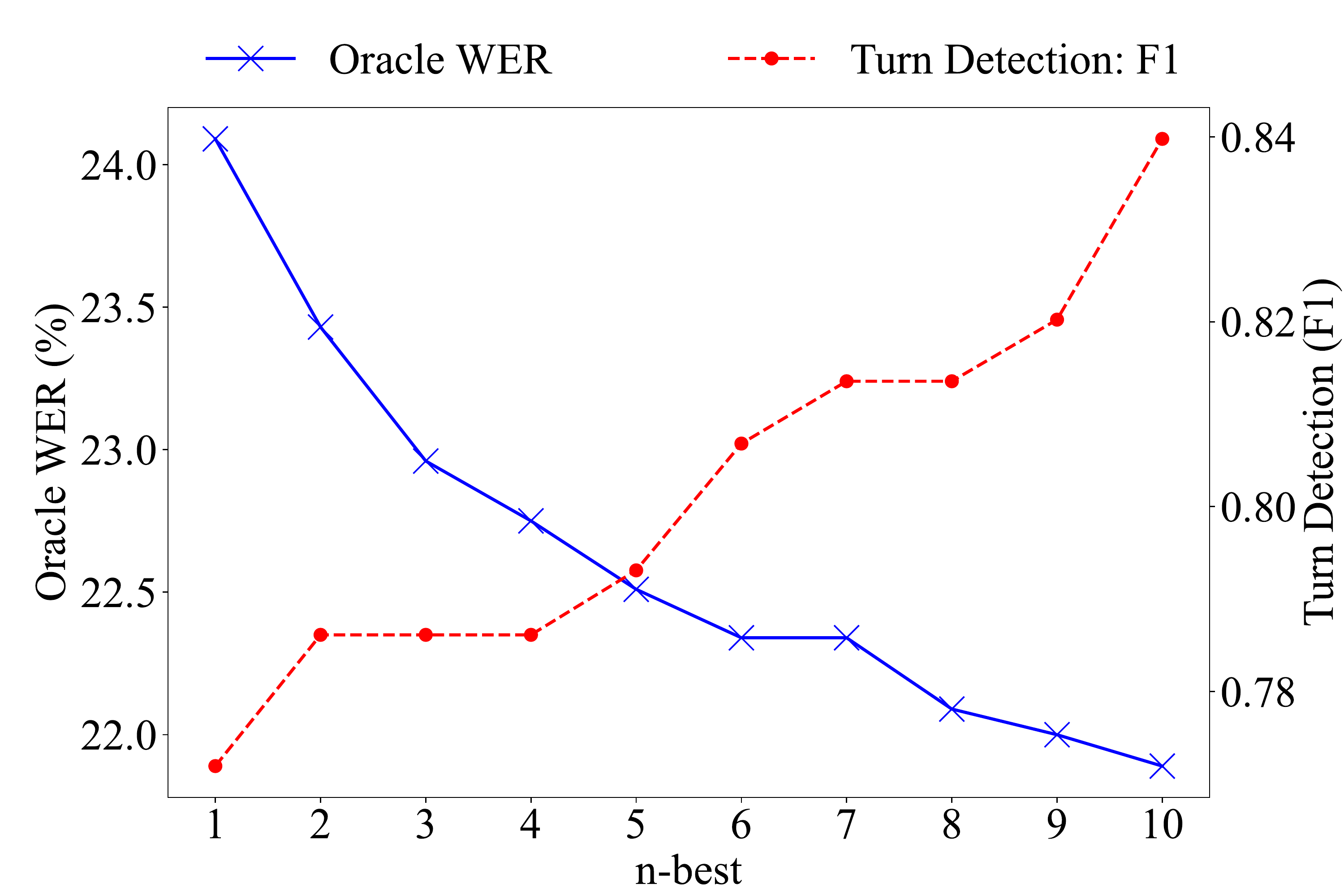}
\caption{Speech recognition and knowledge-seeking turn detection performances with $n$-best ASR hypotheses}
\label{fig:kmdm_nbest}
\end{figure}

To improve the robustness of the baseline model against the ASR errors, especially for the low recall on the detection task,
we developed a simple heuristic to utilize the $n$-best ASR hypotheses.
First, we run the model on each of the $n$-best hypotheses of the target turn.
We still kept the 1-best results for the other previous turns in the dialogue context.
Then, if there is at least one positive prediction from the $n$ runs, we treat the instance as a knowledge-seeking turn.
Figure~\ref{fig:kmdm_nbest} shows that using $n$-best hypotheses with this simple ensemble method helps to improve the detection performance.

\begin{figure}[t]
\centering
\includegraphics[width=\linewidth]{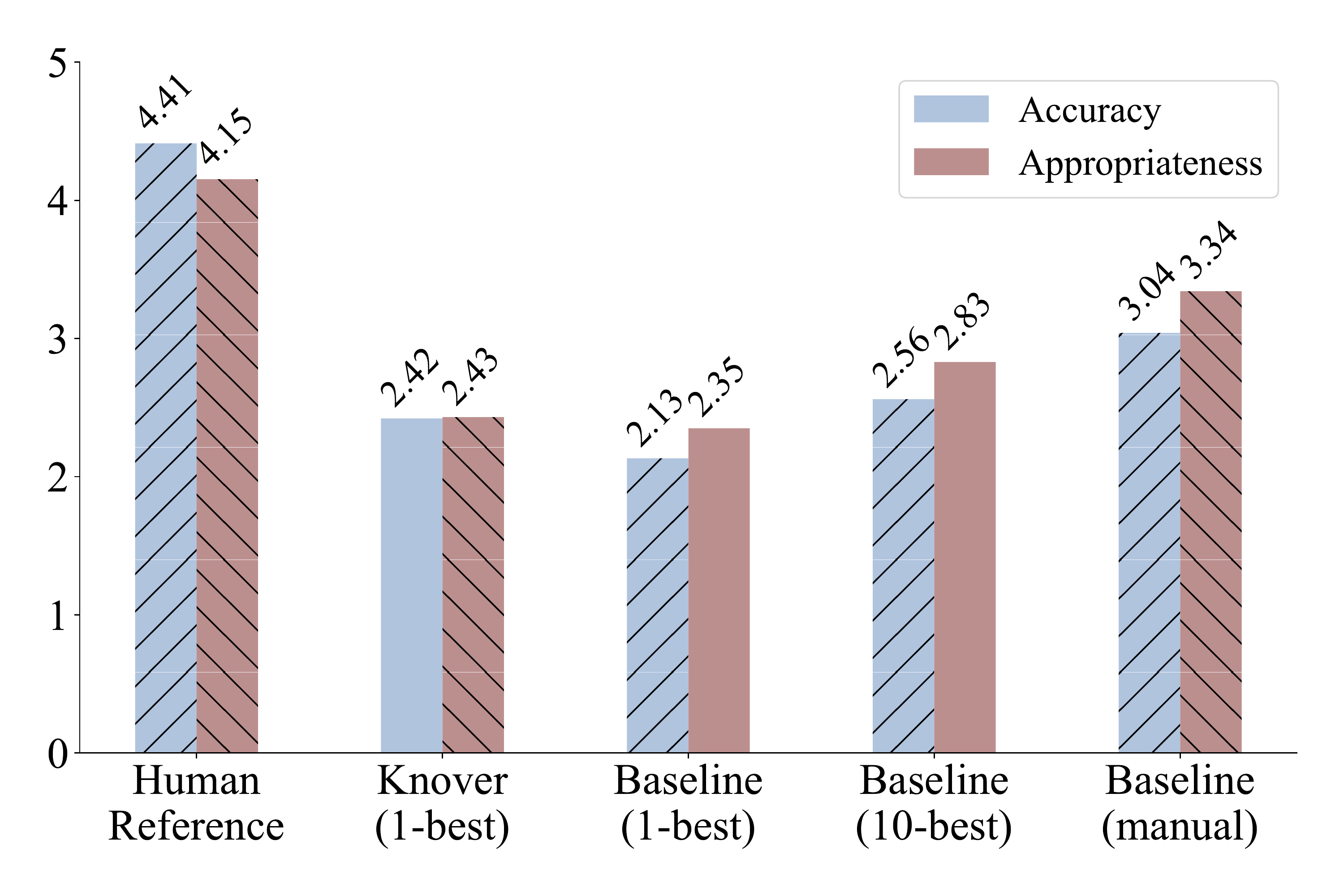}
\caption{Human evaluation results}
\label{fig:human_eval}
\end{figure}

Finally, we conducted a human evaluation with the following crowd-sourcing tasks introduced in~\cite{kim2021domain}:
\begin{itemize}
    \item Accuracy: This task asks crowd workers to score the accuracy of a system output based on the provided reference knowledge on a scale of 1-5.
    \item Appropriateness: This task asks crowd workers to score how well a system output is naturally connected to a given conversation on a scale of 1-5.
\end{itemize}
Figure~\ref{fig:human_eval} compares the human evaluation results of five different configurations in Accuracy and Appropriateness.
As expected, there were still significant differences between all the model outputs and the ground-truth human responses.
Comparing between two base models, Knover was better in both metrics than the DSTC9 baseline, which is consistent with the official DSTC9 results on the written test set.
Again our ensemble heuristic with $n$-best ASR hypotheses helped to boost the baseline performance on the human evaluation results.
Furthermore, the gap between the results on ASR outputs and manual transcripts implies more room for potential improvements with better methods.

\section{Conclusions}
\label{sec:conclusions}

This paper introduced a new data set on multi-domain dialogue state tracking and knowledge-grounded dialogue modeling tasks, aiming towards more robust multi-turn dialogue processing on spoken conversations.
From the evaluations, we observed that the baseline models built on the existing written data were not performing well on the new spoken data for both tasks.
These results show the importance of more speech-oriented studies to improve the robustness of spoken dialogue systems.

To support research in this direction, we organize a public benchmark challenge under DSTC10 and release the data introduced in this work as the official validation set for the challenge participants.
Furthermore, we plan to collect more spoken dialogues and release them as the main challenge test set.
We believe such community efforts will make advancement in the state-of-the-art in spoken dialogue processing.

\bibliographystyle{IEEEbib}
\bibliography{main}

\end{document}